\documentclass{article}

\usepackage{microtype}
\usepackage{graphicx}
\usepackage{subcaption}
\usepackage{booktabs} 

\usepackage[accepted]{icml2019}

\usepackage{color}
\usepackage{caption}
\usepackage{amsfonts}
\usepackage{amsmath}
\usepackage{xspace}
\DeclareMathOperator*{\argmax}{\arg\!\max}

\newcommand{\PolOpt}{\text{\scshape{PolOpt}}\xspace}

\icmltitlerunning{Fingerprint Policy Optimisation for Robust Reinforcement Learning}

\begin{document}

\twocolumn[
\icmltitle{Fingerprint Policy Optimisation \\for Robust Reinforcement Learning}



\icmlsetsymbol{equal}{*}

\begin{icmlauthorlist}
\icmlauthor{Supratik Paul}{cs}
\icmlauthor{Michael A. Osborne}{eng}
\icmlauthor{Shimon Whiteson}{cs}
\end{icmlauthorlist}

\icmlaffiliation{cs}{Department of Computer Science, University of Oxford, UK}
\icmlaffiliation{eng}{Department of Engineering Science, University of Oxford, UK}

\icmlcorrespondingauthor{Supratik Paul}{supratik.paul@cs.ox.ac.uk}

\icmlkeywords{Reinforcement Learning, Bayesian Optimisation, Policy Fingerprint}

\vskip 0.3in
]



\printAffiliationsAndNotice{}  

\begin{abstract}
	Policy gradient methods ignore the potential value of adjusting \emph{environment variables}: unobservable state features that are randomly determined by the environment in a physical setting, but are controllable in a simulator. This can lead to slow learning, or convergence to suboptimal policies, if the environment variable has a large impact on the transition dynamics. In this paper, we present \emph{fingerprint policy optimisation} (FPO), which finds a policy that is optimal in expectation across the distribution of environment variables. The central idea is to use Bayesian optimisation (BO) to actively select the distribution of the environment variable that maximises the improvement generated by each iteration of the policy gradient method. To make this BO practical, we contribute two easy-to-compute low-dimensional \emph{fingerprints} of the current policy. Our experiments show that FPO can efficiently learn policies that are robust to significant rare events, which are unlikely to be observable under random sampling, but are key to learning good policies.
\end{abstract}

\section{Introduction}
Policy gradient methods have demonstrated remarkable success in learning policies for various continuous control tasks \cite{DDPG,Mordatch,GAE}. However, the expense of running physical trials, coupled with the high sample complexity of these methods, pose significant challenges in directly applying them to a physical setting, e.g., to learn a locomotion policy for a robot. Another problem is evaluating the robustness of a learned policy; it is difficult to ensure that the policy performs as expected, as it is usually infeasible to test it across all possible settings. Fortunately, policies can often be trained and tested in a simulator that exposes key \emph{environment variables} -- state features that are unobservable to the agent and randomly determined by the environment in a physical setting, but that are controllable in the simulator. 

Environment variables are ubiquitous in real-world settings.  For example, localisation errors may mean that a robot is much nearer to an obstacle than expected, increasing the risk of a collision.  While an actual robot cannot directly trigger such errors, they are trivial to introduce in simulation.  Similarly, if we want to train a helicopter to fly robustly, we can easily simulate different wind conditions, even though we cannot control weather in the real world.

A na\"{i}ve application of a policy gradient method updates a policy at each iteration by using a batch of trajectories sampled from the original distribution over environment variables irrespective of the current policy or the training iteration. Thus, it does not explicitly take into account how environment variables affect learning for different policies. Furthermore, this approach is not robust to \emph{significant rare events} (SREs), i.e., it fails any time there are rare events that have significantly different rewards thereby affecting expected performance. 

For example, even if robot localisation errors are rare, coping with them properly may still be key to maximising expected performance, since the collisions they can trigger are so catastrophic.  Similarly, handling certain wind conditions properly may be essential to robustly flying a helicopter, even if those conditions are rare, since doing so may avoid a crash.  In such cases, the na\"{i}ve approach will not see such rare events often enough to learn an appropriate response. 

These problems can be avoided by learning \emph{off environment} \cite{frank2008,OFFER,ALOQ}, i.e., exploiting the ability to adjust environment variables in simulation to trigger SREs more often and thereby improve the efficiency and robustness of learning. However, existing off environment approaches have significant disadvantages. They either require substantial prior knowledge about the environment and/or dynamics \cite{frank2008,OFFER}, can only be applied to low dimensional policies \cite{ALOQ}, or are highly sample inefficient \cite{EPOpt}.

In this paper, we propose a new off environment approach called \emph{fingerprint policy optimisation} (FPO) that aims to learn policies that are robust to rare events while addressing the disadvantages mentioned above. At its core, FPO uses a policy gradient method as the policy optimiser. However, unlike the na\"ive approach, FPO explicitly models the effect of the environment variable on the policy updates, as a function of the policy. Using \emph{Bayesian optimisation} (BO), FPO actively selects the environment distribution at each iteration of the policy gradient method in order to maximise the improvement that one policy gradient update step generates. While this can yield biased gradient estimates, FPO implicitly optimises the bias-variance tradeoff in order to maximise its one-step improvement objective.

A key design challenge in FPO is how to represent the current policy, in cases where the policy is a large neural network with thousands of parameters.  To this end, we propose two low-dimensional policy \emph{fingerprints} that act as proxies for the policy.  The first approximates the stationary distribution over states induced by the policy, with a size equal to the dimensionality of the state space.  The second approximates the policy's marginal distribution over actions, with a size equal to the dimensionality of the action space.

We apply FPO to different continuous control tasks and show that it can outperform existing methods, including those for learning in environments with SREs.  We show that both fingerprints work equally well in practice, which implies that, for a given problem, the lower dimensional fingerprint can be chosen without sacrificing performance.

\section{Problem Setting and Background}
\label{sec:background}

A \emph{Markov decision process} (MDP) is a tuple $\langle \mathcal{S}, \mathcal{A}, \mathcal{P}_\theta, r_\theta, s_0, \gamma \rangle$, where $\mathcal{S}$ is the state space, $\mathcal{A}$ the set of actions, $\mathcal{P}$ the transition probabilities, $r$ the reward function, $s_0$ the probability distribution over the initial state, and $\gamma \in [0,1)$ the discount factor. We assume that the transition and reward functions depend on some environmental variables $\theta$. At the beginning of each episode, the environment randomly samples $\theta$ from some (known) distribution $p(\theta)$. The agent's goal is to learn a policy $\pi(a|s)$ mapping states $s \in \mathcal{S}$ to actions $a \in \mathcal{A}$ that maximises the expected return $J(\pi) = \mathbb{E}_{\theta}[R(\theta, \pi)] = \mathbb{E}_{\theta \sim p(\theta), a_t\sim\pi,s_t \sim\mathcal{P} }[\sum_t \gamma^t r_\theta(s_t,a_t)]$. Note that $\pi$ does not condition on $\theta$ at any point. With a slight abuse of notation, we use $\pi$ to denote both the policy and its parameters. We consider environments characterised by significant rare events (SREs), i.e., there exist some low probability values of $\theta$ that generate large magnitude returns (positive or negative), yielding a significant impact on $J(\pi)$.

We assume that learning is undertaken in a simulator, or under laboratory conditions where $\theta$ can be actively set. Simulators are typically imperfect, which can lead to a \emph{reality gap} that inhibits transfer to the real environment. Nonetheless, in most practical settings, e.g., autonomous vehicles and drones, training in the real environment is prohibitively expensive and dangerous, and simulators play an indispensable role.

\vspace{-2mm}
\subsection{Policy Gradient Methods}
\label{sec:pol_grad_methods}

Starting with some policy $\pi_n$ at iteration $n$, gradient based batch policy optimisation methods like REINFORCE \cite{Reinforce}, NPG \cite{NPG}, and TRPO \cite{TRPO} compute an estimate of the gradient $\nabla J(\pi_n)$ by sampling a batch of trajectories from the environment while following $\pi_n$, and then use this estimate to approximate gradient ascent in $J$, yielding an updated policy $\pi_{n+1}$. REINFORCE uses a fixed learning rate to update the policy; NPG and TRPO use the Fisher information matrix to scale the gradient and constrain the KL divergence between consecutive policies, which makes the updates independent of the policy parametrisation.

A major problem for such methods is that the estimate of $\nabla J(\pi)$ can have high variance due to stochasticity in the policy and environment, yielding slow learning \cite{Reinforce,Glynn:1990,PetersRoboticsPG}. In settings with SREs, this problem is compounded by the variance due to $\theta$, which the environment samples for each trajectory in the batch. Furthermore, the SREs may not be observed during learning since the environment samples $\theta \sim p(\theta)$, which can cause convergence to a highly suboptimal policy. Since these methods do not explicitly consider the environment variable's effect on learning, we call them \emph{na\"ive} approaches.

\vspace{-2mm}
\subsection{Bayesian Optimisation}
\label{sec:BO}

A \emph{Gaussian process} (GP) \cite{GPML} is a distribution over functions. It is fully specified by its mean $m(\mathbf{x})$ (often assumed to be 0 for convenience) and covariance functions $k(\mathbf{x}, \mathbf{x'})$ which encode any prior belief about the function. A posterior distribution is generated by using observed values to update the belief about the function in a Bayesian way. The squared exponential kernel is a popular choice for the covariance function, and has the form $k(\mathbf{x}, \mathbf{x'}) = \sigma_0^2 \exp[-\frac{1}{2}(\mathbf{x} - \mathbf{x'})^T\Sigma^{-1} (\mathbf{x} - \mathbf{x'})]$, where $\Sigma$ is a diagonal matrix whose diagonal gives lengthscales corresponding to each dimension of $\mathbf{x}$.
By conditioning on the observed data, predictions for any new points can be computed analytically as a Gaussian $\mathcal{N}\bigl(f(\mathbf{x}); \mu(\mathbf{x}), \sigma^2(\mathbf{x})\bigr)$:
\begin{subequations}
	\label{eq:GP_predictions}
	\begin{align}
	\mu\bigl(\mathbf{x}\bigr) &= k(\mathbf{x}, \mathbf{X})(\mathbf{K}+\sigma^2_{\text{noise}} \mathbf{I})^{-1}f(\mathbf{X}) \\
	\sigma^2\bigl( \mathbf{x} \bigr) &= k(\mathbf{x},\mathbf{x}) 
	- k(\mathbf{x}, \mathbf{X})(\mathbf{K}+\sigma^2_{\text{noise}} \mathbf{I})^{-1}k(\mathbf{X}, \mathbf{x}),
	\end{align}
\end{subequations}
where $\mathbf{X}$ is the design matrix, $f(\mathbf{X})$ is the corresponding function values, and $\mathbf{K}$ is the covariance matrix with elements $k(\mathbf{x_i}, \mathbf{x_j})$. Probabilistic modelling of the predictions makes GPs well suited for optimising $f(\mathbf{x})$ using BO. Given a set of observations, the next point for evaluation is chosen as the $\mathbf{x}$ that maximises an \emph{acquisition function}, which uses the posterior mean and variance to balance exploitation and exploration. The choice of acquisition function can significantly impact performance and numerous acquisition functions have been suggested in the literature, ranging from confidence or expectation based methods to entropy based methods. We consider two acquisition functions: \emph{upper confidence bound} (UCB) \cite{cox92sdo,cox97sdo}, and \emph{fast information-theoretic Bayesian optimisation} (FITBO) \cite{FITBO}. 

Given a dataset $\mathcal{D}_{1:n} = \{(\mathbf{x}_i, f(\mathbf{x}_i))\}_{i=1}^n$, UCB directly incorporates the prediction uncertainty by defining an upper bound: $\alpha_{\text{UCB}}(\mathbf{x}\mid \mathcal{D}_{1:n}) = \mu(f(\mathbf{x})\mid \mathcal{D}_{1:n}) + \kappa\, \sigma(f(\mathbf{x})\mid \mathcal{D}_{1:n})$, where $\kappa$ controls the exploration-exploitation tradeoff. By contrast,  FITBO aims to reduce the uncertainty about the global optimum $f(\mathbf{x}^*)$ by selecting the $\mathbf{x}$ that minimises the entropy of the distribution $p(f(\mathbf{x}^*)\mid \mathbf{x}, \mathcal{D}_{1:n})$:
\begin{align}
& \alpha_{\text{FITBO}}(\mathbf{x}\mid \mathcal{D}_{1:n}) = H[p(f(\mathbf{x})\mid \mathbf{x}, \mathcal{D}_{1:n})] \nonumber \\
& - \mathbb{E}_{p(f(\mathbf{x}^*)\mid \mathcal{D}_{1:n})} \bigl[H[p(f(\mathbf{x})\mid f(\mathbf{x}^*), \mathbf{x}, \mathcal{D}_{1:n})]\bigr] \nonumber,
\end{align}
which cannot be computed analytically, but can be approximated efficiently following  \citet{FITBO}.

BO mainly minimises simple regret: The acquisition function suggests the next point $\mathbf{x}_i$ for evaluation at each timestep, but then the algorithm suggests what it believes to be the optimal point $\hat{\mathbf{x}}^*_i$, and the regret is defined as $\sum_{i=1}^N f(\mathbf{x}^*) - f(\hat{\mathbf{x}}^*_i)$. This is different from a bandit setting where the cumulative regret is defined as $\sum_{i=1}^N f(\mathbf{x}^*) - f(\mathbf{x}_i)$. \citet{krause2011contextual} show that the UCB acquisition function is also a viable strategy to minimise cumulative regret in a contextual GP bandit setting, where selection of $\mathbf{x}_i$ conditions on some observed context.

\begin{figure*}[t]
	\centering
	\begin{subfigure}{0.32\linewidth}
		\centering
		\includegraphics[width=1\linewidth]{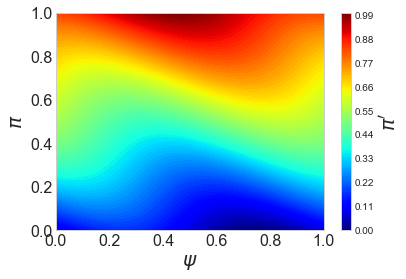}
		\caption{}
	\end{subfigure}
	\begin{subfigure}{0.32\linewidth}
		\centering
		\includegraphics[width=1\linewidth]{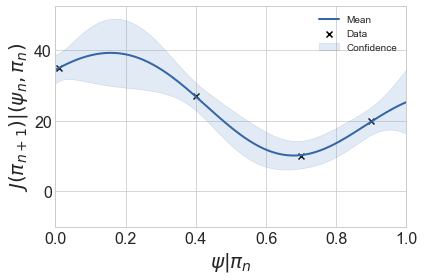}
		\caption{}
		\label{fig:FPO_b}
	\end{subfigure}
	\begin{subfigure}{0.32\linewidth}
		\centering
		\includegraphics[width=1\linewidth]{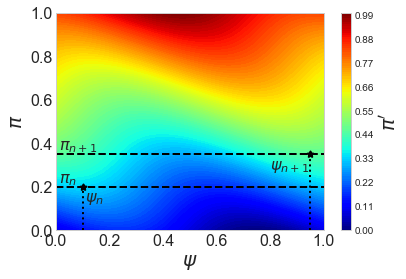}
		\caption{}
	\end{subfigure}
	\vspace{-2mm}
	\caption{(a) The policy optimisation routine \PolOpt takes input $(\psi,\pi)$ and updates the policy to $\pi'$; (b) FPO directly models $J(\pi')$ as a function of $(\psi, \pi)$. At iteration $n$, $\pi_n$ is fixed and FPO selects $\psi_n|\pi_n$ based on the UCB or FITBO acquisition function, (c) after which $\PolOpt(\psi_n, \pi_n)$ updates the policy to $\pi_{n+1}$, and this process repeats.} 
	\label{fig:FPO}
	\vspace{-1mm}
\end{figure*}

\section{Fingerprint Policy Optimisation}
\label{sec:FPO}

To address the challenges posed by environments with SREs, we introduce \emph{fingerprint policy optimisation} (FPO). The main idea is to sample $\theta$ from a parametrised distribution $q_{\psi_n}(\theta)$ where the parameters $\psi_n$ are conditioned on a fingerprint of the current policy, such that it helps the policy optimisation routine learn a policy that takes into account any SREs. Concretely, FPO executes the following steps at each iteration $n$.  First, it selects $\psi_n$ by approximately solving an optimisation problem defined below.  Second, it samples trajectories from the environment using the current policy $\pi_n$, where each trajectory uses a value for $\theta$ sampled from a distribution parameterised by $\psi_n$: $\theta \sim q_{\psi_n}(\theta)$.  Third, these trajectories are fed to a policy optimisation routine, e.g., a policy gradient algorithm, which uses them to compute an updated policy, $\pi_{n+1} = \PolOpt(\psi, \pi_n)$.  Fourth, new, independent trajectories are generated from the environment with $\pi_{n+1}$ and used to estimate $J(\pi_{n+1})$.  Fifth, a new point is added to a dataset $\mathcal{D}_{1:n} = \{\bigl((\psi_{i-1}, \pi_{i-1}), J(\pi_i)\bigr)\}_{i=1}^n$, which is input to a GP.  The process then repeats, with BO using the GP to select the next $\psi$.

The key insight behind FPO is that at each iteration $n$, FPO should select the $\psi_n$ that it expects will maximise the performance of the \emph{next} policy, $\pi_{n+1}$:
\begin{align}
\label{eq:obj_psi}
\psi_n|\pi_n &= \argmax_\psi J(\pi_{n+1}) \nonumber \\
&= \argmax_\psi J \bigl(\PolOpt(\psi, \pi_n)\bigr).
\end{align}

In other words, FPO chooses the $\psi_n$ that it thinks will help \PolOpt maximise the improvement to $\pi_n$ that can be made in a single policy update.  By modelling the relationship between $\psi_n$, $\pi_n$, and $J(\pi_{n+1})$ with a GP, FPO can learn from experience how to select an appropriate $\psi$ for the current $\pi$. Modelling $J(\pi_{n+1})$ directly also bypasses the issue of modelling $\pi_{n+1}$ and its relationship to $J(\pi_{n+1})$, which is infeasible when $\pi_{n+1}$ is high dimensional. Note that while \PolOpt has inputs $(\psi,\pi_n)$, the optimisation is performed over $\psi$ only, with $\pi_n$ fixed. FPO is summarised in Algorithm \ref{algo:CB-PolOpt} and illustrated in Figure \ref{fig:FPO}. The remainder of this section describes in more detail elements of FPO that are essential to make it work in practice.

\vspace{-2mm}
\subsection{Selecting $\psi_n$}
The optimisation problem in \eqref{eq:obj_psi} is difficult for two reasons.  First, solving it requires calling \PolOpt, which is  expensive in both computation and samples. Second, the observed $J(\pi)$ can be noisy due to the inherent stochasticity in the policy and the environment.

BO is particularly suited to such settings as it is sample efficient, gradient free, and can work with noisy observations. In this paper, we consider both the UCB and FITBO acquisition functions to select $\psi_n|\pi_n$ in \eqref{eq:obj_psi}, and compare their performance. Formally, we model the returns $J(\pi_n)$ as a GP with inputs $(\psi_{n-1}, \pi_{n-1})$. Given a dataset $\mathcal{D}_{1:n} = \{((\psi_{i-1}, \pi_{i-1}), J(\pi_i))\}_{i=1}^n$, $\psi_n$ is selected by maximising the UCB or FITBO acquisition function:
\begin{align}
\label{eq:ucb_acqfn}
&\alpha_{\text{UCB}}(\psi_n|\pi_n) = \mu(J(\pi_{n+1})|\pi_n) + \kappa \sigma(J(\pi_{n+1})|\pi_n)\\
\label{eq:fitbo_acqfn}
&\alpha_{\text{FITBO}}(\psi_n|\pi_n) = H[p(J(\pi_{n+1})|\psi, \pi_n)] \nonumber \\
&- \mathbb{E}_{p(J(\pi_{n+1}^*)|\pi_n)} [H[p(J(\pi_{n+1})|\psi, \pi_n, J(\pi_{n+1}^*))]],
\end{align}
where we drop conditioning on $\mathcal{D}_{1:n}$ for ease of notation. See Figure \ref{fig:FPO_b} for an illustration.

Estimating the gradient using trajectories sampled from the environment with $\theta \sim q_\psi(\theta)$ introduces bias. While importance sampling methods  \citep{frank2008,OFFER} could correct for this bias, FPO does not explicitly do so. Instead FPO lets BO implicitly optimise a bias-variance tradeoff by selecting $\psi$ to maximise the one-step improvement objective.

\begin{algorithm}[t]
	\caption{Fingerprint Policy Optimisation}
	\label{algo:CB-PolOpt}
	\begin{algorithmic}[1]
		\INPUT Initial policy $\pi_0$, original distribution $p(\theta)$, randomly initialised $q_{\psi_0}(\theta)$, policy optimisation method \PolOpt, number of policy iterations $N$, dataset $\mathcal{D}_0 = \{\}$
		\FOR {n = 1, 2, 3, \ldots, N}
		\STATE Sample $\theta_{1:k}$ from $q_{\psi_{n-1}}(\theta)$, and with $\pi_{n-1}$ sample trajectories $\tau_{1:k}$ corresponding to each $\theta_{1:k}$
		\STATE Compute $\pi_n = \PolOpt(\tau_{1:k}) = \PolOpt(\psi_{n-1}, \pi_{n-1})$
		\STATE Compute $J(\pi_n)$ using numerical quadrature as described in Section \ref{sec:estimate_pol_val}. Use the sampled trajectories to compute the policy fingerprint as described in Section \ref{sec:pol_fingerprint}.
		\STATE Set $\mathcal{D}_n = \mathcal{D}_{n-1} \cup \{((\psi_{n-1}, \pi_{n-1}), J(\pi_n)) \}$ and update the GP to condition on $\mathcal{D}_n$
		\STATE Use either the UCB \eqref{eq:ucb_acqfn} or FITBO \eqref{eq:fitbo_acqfn} acquisition functions to select $\psi_n$.
		\ENDFOR
	\end{algorithmic}
\end{algorithm}

\subsection{Estimating $J(\pi_n)$}
\label{sec:estimate_pol_val}

Estimating $J(\pi_n)$ accurately in the presence of SREs can be challenging. A Monte Carlo estimate using samples of trajectories from the original environment requires prohibitively many samples. One alternative would be to apply an IS correction to the trajectories generated from $q_{\psi_n}(\theta)$ for the policy optimisation routine. However, this is not feasible since it would require computing the IS weights  $\frac{p(\tau|\theta)}{q_\psi(\tau|\theta)}$, which depend on the unknown transition function. Furthermore, even if the transition function is known, there is no reason why $\psi_n$ should yield a good IS distribution since it is selected with the objective of maximising $J(\pi_{n+1})$.

Instead, FPO applies exhaustive summation for discrete $\theta$ and numerical quadrature for continuous $\theta$ to estimate $J(\pi_n)$. That is, if the support of $\theta$ is discrete, FPO simply samples a trajectory from each environment defined by $\theta$ and estimates $J(\pi_n) = \sum_{l=1}^{L} p(\theta=\theta_l)R(\theta_l, \pi_n)$. To reduce the variance due to stochasticity in the policy and the environment, we can sample multiple trajectories from each $\theta_l$. For continuous $\theta$, we apply an adaptive Gauss-Kronrod quadrature rule to estimate $J(\pi_n)$.  While numerical quadrature may not scale to high dimensions, in practice $\theta$ is usually low dimensional, making this a practical design choice.

Since for discrete $\theta$ we evaluate $J(\pi_n)$ through exhaustive summation, it is natural to consider a variation of the na\"ive approach, wherein $\nabla J(\pi_n)$ is also evaluated in the same manner during training, i.e., $\nabla J(\pi_n) = \sum_{l=1}^{L} p(\theta=\theta_l)\nabla J(\theta_l, \pi_n)$. We call this the `Enum' baseline since the gradient is estimated by enumerating over all possible values of $\theta$. Our experiments in Section \ref{sec:experiments} show that this baseline is unable to match the performance of FPO.

\subsection{Policy Fingerprints}
\label{sec:pol_fingerprint}

True global optimisation is limited by the curse of dimensionality to low-dimensional inputs, and BO has had only rare successes in problems with more than twenty dimensions \cite{wang_bayesian_2013}. In FPO, many of the inputs to the GP are policy parameters and in practice, the policy may be a neural network with thousands of parameters. While \citet{linear_policies} show that linear and radial basis function policies can perform as well as neural networks in some simulated continuous control tasks, even these policies have hundreds of parameters, far too many for a GP.

Thus, we need to develop a policy \emph{fingerprint}, i.e., a representation that is low dimensional enough to be treated as an input to the GP but expressive enough to distinguish the policy from others. \citet{StabExpReplay} showed that a surprisingly simple fingerprint, consisting only of the training iteration, suffices to stabilise multi-agent $Q$-learning.  Such a fingerprint is insufficient for FPO, as the GP fails to model the response surface and treats all observed $J(\pi)$ as noise. However, the principle still applies: a simplistic fingerprint that discards much information about the policy can still be sufficient for decision making, in this case to select $\psi_n$.

In this spirit, we propose two fingerprints.  The first, the \emph{state fingerprint}, augments the training iteration with an estimate of the stationary state distribution induced by the policy.  In particular, we fit an anisotropic Gaussian to the set of states visited in the trajectories sampled while estimating $J(\pi)$ (see Section \ref{sec:estimate_pol_val}).  The size of this fingerprint grows linearly with the dimensionality of the state space, instead of the number of parameters in the policy.

In many settings, the state space is high dimensional, but the action space is low dimensional. Therefore, our second fingerprint, the \emph{action fingerprint}, is a Gaussian approximation of the marginal distribution over actions induced by the policy: $a(\pi) = \int \pi(a|s)s(\pi)ds$ (here $s(\pi)$ is the stationary state distribution induced by $\pi$), sampled from trajectories as with the state fingerprint.

Of course, neither the stationary state distribution nor the marginal action distribution are likely to be Gaussian and could in fact be multimodal.  Furthermore, the state distribution is estimated  from samples used to estimate $J(\pi)$, and not from $p(\theta)$. However, as our results show, these representations are nonetheless effective, as they do not need to accurately describe each policy, but instead just serve as low dimensional fingerprints on which FPO conditions.

\vspace{-2mm}
\subsection{Covariance Function}
Our choice of policy fingerprints means that one of the inputs to the GP is a probability distribution. Thus for our GP prior we use a covariance that is the product of three terms, $k_1(\psi, \psi'), k_2(n,n')$ and $k_3\bigl(\textsc{fgp}(\pi), \textsc{fgp}(\pi')\bigr)$, where each of $k_1, k_2$, and $k_3$ is a squared exponential covariance function and $\textsc{fgp}(\pi)$ is the state or action fingerprint of $\pi$. Similar to \citet{Hellinger_kernel}, we use the Hellinger distance to replace the Euclidean in $k_3$: this covariance remains positive-semi-definite as the Hellinger is effectively a modified Euclidean.

\section{Related Work}
\label{sec:related_work}

\begin{figure*}[t]
	\centering
	\begin{subfigure}{0.32\linewidth}
		\includegraphics[width=1\linewidth]{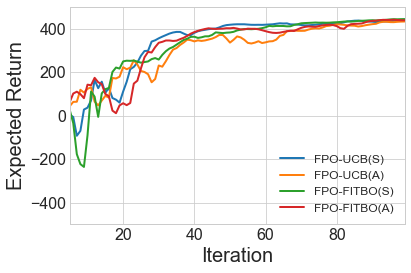}
		\caption{Different versions of FPO}
		\label{fig:cliff_q1}
	\end{subfigure}\hfill
	\begin{subfigure}{0.32\linewidth}
		\includegraphics[width=1\linewidth]{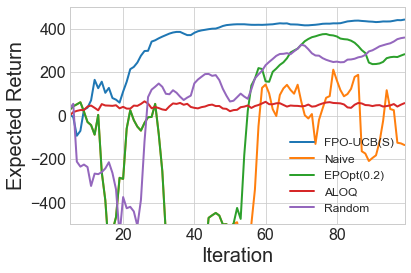}
		\caption{FPO vs baselines}
		\label{fig:cliff_q2}
	\end{subfigure}
	\begin{subfigure}{0.32\linewidth}
		\includegraphics[width=1\linewidth]{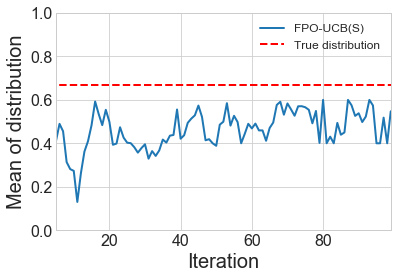}
		\caption{FPO vs baselines}
		\label{fig:cliff_mean_theta}
	\end{subfigure}
	\vspace{-3mm}
	\caption{Results for the Cliff Walker environment. (a) Comparison of the performance of different versions of FPO. (b) Comparison of FPO-UCB(S) against the baselines. (c) The mean of the true distribution of the cliff location (`True distribution'), and the mean of the distribution selected by FPO-UCB(S) during learning (FPO-UCB(S)). Information on variation in performance due to the random starts is presented in Appendix \ref{app:quartiles}.}
	\vspace{-4mm}
\end{figure*}

Various methods have been proposed for learning in the presence of SREs. These are usually off environment and either based on learning a good IS distribution from which to sample the environment variable \cite{frank2008,OFFER}, or Bayesian active selection of the environment variable during learning \cite{ALOQ}.

\citet{frank2008} propose a temporal difference based method that uses IS to efficiently evaluate policies whose expected value may be substantially affected by rare events.  However, their method assumes prior knowledge of the SREs, such that they can directly alter the probability of such events during policy evaluation. By contrast, FPO does not require any such prior knowledge about SREs, or the environment variable settings that might trigger them. It only assumes that the original distribution of the environment variable is known, and that the environment variable is controllable during learning.

OFFER \cite{OFFER} is a policy gradient method that uses observed trials to gradually change the IS distribution over the environment variable. Like FPO, it makes no prior assumptions about SREs. However, at each iteration it updates the environment distribution with the objective of minimising the variance of the gradient estimate, which may not lead to the distribution that optimises the learning of the policy. Furthermore, OFFER requires a full transition model of the environment to compute the IS weights. It can also lead to unstable IS estimates if the environment variable affects any transitions besides the initial state.

ALOQ \cite{ALOQ} is a Bayesian optimisation and quadrature method that models the return as a GP with the policy parameters and environment variable as inputs. At each iteration it actively selects the policy and then the environment variable in an alternating fashion and, as such, performs the policy search in the parameter space. As a BO based method, it does not assume the observations are Markov and is highly sample efficient. However, it can only be applied to settings with low dimensional policies. Furthermore, its computational cost scales cubically with the number of iterations, and is thus limited to settings where a good policy can be found within relatively few iterations. By contrast, FPO uses policy gradients to perform policy optimisation while the BO component generates trajectories, which when used by the policy optimiser are expected to lead to a larger improvement in the policy. 

In the wider BO literature, \citet{williams_santner} suggested a method for settings where the objective is to optimise expensive integrands. However, their method does not specifically consider the impact of SREs and, as shown by \citet{ALOQ}, is unsuitable for such settings. \citet{BO_expensive_integrands} suggest BQO, another BO based method for expensive integrands. Their method also does not explicitly consider SREs. Finally, both these methods suffer from all the disadvantages of BO based methods mentioned earlier.

EPOpt($\epsilon$) \citep{EPOpt} is a different off-environment approach that learns robust policies by maximising the $\epsilon$-percentile conditional value at risk (CVaR) of the policy. First, it randomly samples a set of simulator settings; then trajectories are sampled for each of these settings. A policy optimisation routine (e.g., TRPO \cite{TRPO}) then updates the policy based on only those trajectories with returns lower than the $\epsilon$ percentile in the batch. A fundamental difference to FPO is that it finds a risk-averse solution based on CVaR, while FPO finds a risk neutral policy. Also, while FPO actively changes the distribution for sampling the environment variable at each iteration, EPOpt samples them from the original distribution, and is thus unlikely to be suitable for settings with SREs, since it will not generate SREs often enough to to learn an appropriate response. Finally, EPOpt discards all sampled trajectories in the batch with returns greater than $\epsilon$ percentile for use by the policy optimisation routine, making it highly sample inefficient, especially for low values of $\epsilon$.

RARL \citep{RARL} learns robust policies by training in a simulator where an adversary applies destabilising forces, with both the agent and the adversary trained simultaneously. RARL requires significant prior knowledge in setting up the adversary to ensure that the environment is easy enough to be learnable but difficult enough that the learned policy will be robust.  Like EPOpt, it does not consider settings with SREs.

By learning an optimal distribution for the environment variable conditioned on the policy fingerprint, FPO also has some parallels with meta-learning. Methods like MAML \cite{MAML}, and Reptile \cite{Reptile} seek to find a good policy representation that can be adapted quickly to a specified task. \citet{learn_by_grad,learn_without_grad} seek to optimise neural networks by learning an automatic update rule based on transferring knowledge from similar optimisation problems. To maximise the performance of a neural network across a set of discrete tasks, \citet{Graves} propose a method for automatically selecting a curriculum during learning. Their method treats the problem as a multi-armed bandit and uses Exp3 \cite{Exp3} to find the optimal curriculum. Unlike these methods, which seek to quickly adapt to a new task after training on some related task, FPO seeks to maximise the expected return across a family of tasks characterised by different settings of the environment variable.

\begin{figure*}[t]
	\centering
	\begin{subfigure}{0.32\linewidth}
		\includegraphics[width=1\linewidth]{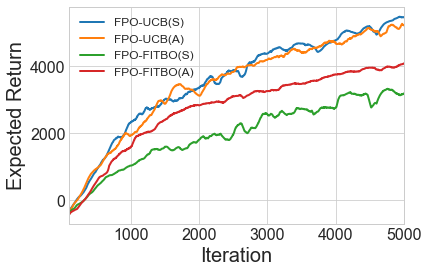}
		\caption{Different versions of FPO}
		\label{fig:cheetah_q1}
	\end{subfigure}\hfill
	\begin{subfigure}{0.32\linewidth}
		\includegraphics[width=1\linewidth]{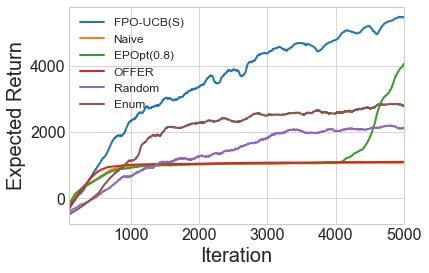}
		\caption{FPO vs baselines}
		\label{fig:cheetah_q2}
	\end{subfigure}\hfill
	\begin{subfigure}{0.32\linewidth}
		\includegraphics[width=1\linewidth]{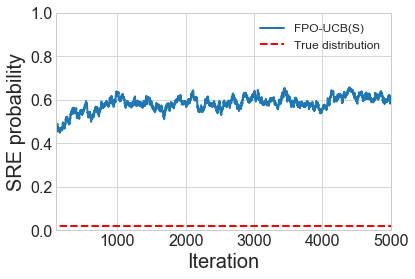}
		\caption{Learnt $\psi$}
		\label{fig:cheetah_psi}
	\end{subfigure}\hfill
	\vspace{-3mm}
	\caption{Results for the Half Cheetah environment. (a) Comparison of the performance of different versions of FPO. (b) Comparison of FPO-UCB(S) against the baselines. (c) The actual probability of velocity target being 4 (`True distribution'), and the probability selected by FPO-UCB(S) during learning (FPO-UCB(S)). Information on variation in performance due to the random starts is presented in Appendix \ref{app:quartiles}.}
\end{figure*}

\section{Experiments}
\label{sec:experiments}
To evaluate the empirical performance of FPO, we start by applying it to a simple problem: a modified version of the cliff walker task \cite{Sutton&Barto}, with one dimensional state and action spaces. We then move on to simulated robotics problems based on the MuJoCo simulator \cite{OpenAIGym} with much higher dimensionalities. These were modified to include SREs. We aim to answer two questions: (1) How do the different versions of FPO (UCB vs.\ FITBO acquisition functions, state (S) vs.\ action (A) fingerprints) compare with each other? (2) How does FPO compare to existing methods (Na\"ive, Enum, OFFER, EPOpt, ALOQ), and ablated versions of FPO.  We use TRPO as the policy optimisation method combined with neural net policies. We also include $\theta$ as an input to the baseline (but not to the policy) since it is observable during training.

We repeat all our experiments across 10 random starts. For ease of viewing we present only the median of the expected returns in the plots. Further experimental details are provided in Appendix \ref{app:hyper_settings}, and information about the variation due to the random starts is presented in Appendix \ref{app:quartiles}. 

Due to the disadvantages of ALOQ mentioned in Section \ref{sec:related_work}, we were able to apply it only on the cliff walker problem.  The policy dimensionality and the total number of iterations for the simulated robotic tasks were far too high. Note also that, while we compare FPO to EPOpt, these methods optimise for different objectives.

\begin{figure*}
	\centering
	\begin{subfigure}{0.32\linewidth}
		\includegraphics[width=1\linewidth]{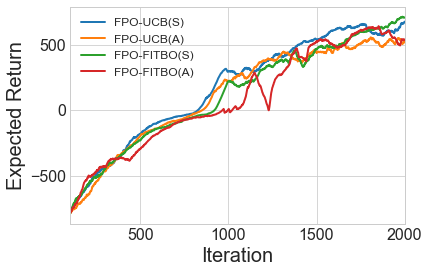}
		\caption{Different versions of FPO}
		\label{fig:ant_q1}
	\end{subfigure}\hfill
	\begin{subfigure}{0.32\linewidth}
		\includegraphics[width=1\linewidth]{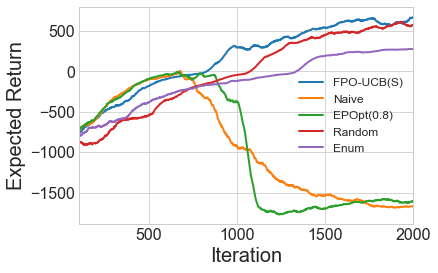}
		\caption{FPO vs baselines}
		\label{fig:ant_q2}
	\end{subfigure}\hfill
	\begin{subfigure}{0.32\linewidth}
		\includegraphics[width=1\linewidth]{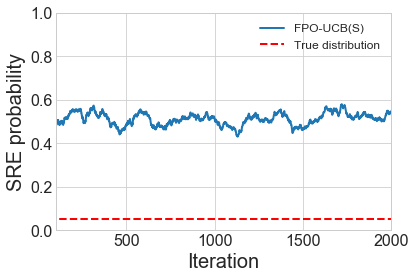}
		\caption{Learnt $\psi$}
		\label{fig:ant_psi}
	\end{subfigure}
	\vspace{-3mm}
	\caption{Results for the Ant environment. (a) Comparison of the performance of different versions of FPO. (b) Comparison of FPO-UCB(S) against the baselines. (c) The actual probability of velocity target being 4 (`True distribution'), and the probability selected by FPO-UCB(S) during learning (FPO-UCB(S)). Information on variation in performance due to the random starts is presented in Appendix \ref{app:quartiles}.}
	\vspace{-4mm}
\end{figure*}

\vspace{-1mm}
\subsection{Cliff Walker}
\label{sec:exp_cliff_walker}
\vspace{-1mm}
We start with a modified version of the cliff walker problem where instead of a gridworld we consider an agent moving in a continuous state space $\mathbb{R}$; the agent starts randomly near the state 0, and at each timestep can take an action $a \in \mathbb{R}$. The environment then transitions the agent to a new location $s' = s + 0.025\text{sign}(a) + 0.005\epsilon$, where $\epsilon$ is standard Gaussian noise. The location of the cliff is given by $1+\theta$, where $p(\theta) \sim \textit{Beta}(2,1)$. If the agent's current state is lower than the cliff location, it gets a reward equal to its state; otherwise it falls off the cliff and gets a reward of -5000, terminating the episode. Thus the objective is to learn to walk as close to the cliff edge as possible without falling over.

Figure \ref{fig:cliff_q1} shows that all versions of FPO do equally well on the task. Figure \ref{fig:cliff_q2} shows that FPO-UCB(S) learns a policy with a higher expected return than all other baselines. This is not surprising since, as discussed in Section \ref{sec:pol_grad_methods}, the gradient estimates without active selection of $\psi$ are likely to have high variance due to the presence of SREs. For EPOpt we set $\epsilon=0.2$ and perform rejection sampling after 50 iterations. The poor performance of ALOQ is expected since even in this simple problem, the policy dimensionality of 47 is high for BO. We could not run OFFER since an analytical solution of $\nabla_\psi q_\psi(\theta)$ does not exist.

Figure \ref{fig:cliff_mean_theta} compares the mean of sampling distribution for the cliff location chosen by FPO-UCB(S), i.e., $\mathbb{E}_{q_{\psi_n}} [\theta]$, against the mean of true distribution of the cliff location ($\mathbb{E}_{p}[\theta]$). FPO-UCB(S) selects sampling distributions for the cliff location that are more conservative than the true distribution. This is expected as falling off the cliff has a significant cost  and closer cliff locations need to be sampled more frequently to learn a policy that avoids the low probability events where the cliff location is close under the true distribution.

The large negative reward for falling off the cliff acts like an SRE in this setting, so we can modify it to have no SRE by setting the reward to 0, but retain the randomised location of the cliff as an environment variable that affects the dynamics. In Appendix \ref{app:no_sre_cliff} we show that the active selection of the environment variable enables FPO to outperform the na\"ive approach in this setting as well.

\vspace{-1mm}
\subsection{Half Cheetah}
\vspace{-1mm}
Next we consider simulated robotic locomotion tasks using the Mujoco simulator. In the original OpenAI Gym HalfCheetah task, the objective is to maximise forward velocity. We modify the original problem such that in 98\% of the cases the objective is to achieve a target velocity of 2, with rewards decreasing linearly with the distance from the target. In the remaining 2\%, the target velocity is set to 4, with a large bonus reward, which acts as an SRE.

Figure \ref{fig:cheetah_q1} shows that FPO with UCB outperforms FPO with FITBO. We suspect that this is because FITBO tends to over-explore. Also, as mentioned in Section \ref{sec:BO}, it was developed with the aim of minimising simple regret and it is not known how efficient it is at minimising cumulative regret. As in cliff walker, both the action and state fingerprints perform equally well with UCB.

Figure \ref{fig:cheetah_q2} shows that the Na\"ive method and OFFER converge to a locally optimal policy that completely ignores the SRE, while the random selection of $\psi$ does slightly better. The Enum baseline performs better than random, but is still far worse than FPO. This shows that the key to learning a good policy is the active selection of $\psi_n$, which also includes the implicit bias-variance tradeoff performed by FPO. We set $\epsilon=0.8$ for EPOpt, but in this case we use the trajectories with returns exceeding the threshold for the policy optimisation since the SRE has a large positive return. Although its performance increases after iteration 4,000, it is extremely sample inefficient, requiring about five times the samples of FPO. We did not run ALOQ as it is entirely infeasible given the policy dimensionality ($>20,000$).

Finally, Figure \ref{fig:cheetah_psi} presents the schedule of $\psi$, (i.e., the probability of the target velocity being 4) as selected by FPO-UCB(S) across different iterations. FPO learns to vary $\psi$, starting with 0.5 initially, to hovering around 0.6 once the cheetah can consistently reach velocities greater than 2.  Further assessments of the learnt policy is presented in Appendix \ref{app:more_experiments}.

\vspace{-1mm}
\subsection{Ant}
\vspace{-1mm}
The ant environment is much more difficult than half cheetah since the agent moves in 3D, and the state space has 111 dimensions compared to 17 for half cheetah. The larger state space also makes learning difficult due to the higher variance in the gradient estimates. We modify the original problem such that velocities greater than 2 carry a 5\% chance of damage to the ant. On incurring damage, which we treat as the SRE, the agent receives a large negative reward, and the episode terminates.

Figure \ref{fig:ant_q1} compares the performance of the UCB versions of FPO.  There is no significant difference between the performance of the state or action fingerprint, or between the UCB and FITBO acquisition functions. Thus, the lower dimensional fingerprint (in this case the action fingerprint) can be chosen without compromising performance.

Figure \ref{fig:ant_q2} shows that for the Na\"ive method and EPOpt, performances drops significantly after about 750 iterations. This is because the learned policies generate velocities beyond 2, which after factoring in the effect of the SRE, yield much lower expected returns. Since the SREs are not seen often enough, these methods do not learn that higher velocities actually yield lower expected returns. The Enum baseline once again performs better than the na\"ive approach, but still worse than FPO. The random baseline performs better, and eventually matches the performance of FPO. We could not run OFFER in this setting as computing the IS weights would require knowledge of the transition model.

Figure \ref{fig:ant_psi} shows that the optimal schedule for $\psi$, the probability of damage for velocities greater than 2, as learnt by FPO-UCB(S), hovers around 0.5. This helps explain the relatively good performance of the random baseline: since the baseline samples $\psi \sim U(0,1)$ at each iteration, the expected value of $\psi$ is 0.5, and thus we can expect it to find a good policy eventually. Of course, active selection of $\psi$ still matters, as evidenced by FPO outperforming it initially. 

\begin{figure}
	\centering
	\includegraphics[width=0.9\linewidth]{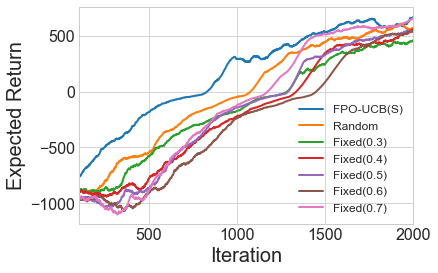}
	\vspace{-2mm}
	\caption{Comparison of FPO-UCB(S) against the Fixed baselines}
	\label{fig:ant_bl}
	\vspace{-5mm}
\end{figure}

A natural question that arises based on the above result, is whether the schedule learnt by FPO-UCB(S) is better than a fixed schedule wherein the probability of breakage is fixed to some value throughout training. We investigate this by running another set of baselines wherein the probability of breakage is fixed to each of $\{0.3,0,4,0.5,0.6,0.7\}$ throughout training. The learning curves are presented in Figure \ref{fig:ant_bl}, with `Fixed$(x)$'  referring to the baseline with the probability of breakage fixed at $x$. While all of the baselines finally converge to a policy that performs on par with FPO-UCB(S), they learn much slower initially. This goes to show that learning a schedule that takes into account the current policy is key to faster learning. Further assessments of the learnt policy is presented in Appendix \ref{app:more_experiments}.

\vspace{-2mm}
\section{Conclusions \& Future Work}
\vspace{-1mm}
Environment variables can have a significant impact on the performance of a policy and are pervasive in real-world settings. This paper presented FPO, a method based on the insight that active selection of environment variables during learning can lead to policies that take into account the effect of SREs. We introduced novel state and action fingerprints that can be used by BO with a one-step improvement objective to make FPO scalable to high dimensional tasks irrespective of the policy dimensionality. We applied FPO to a number of continuous control tasks of varying difficulty and showed that FPO can efficiently learn policies that are robust to significant rare events, which are unlikely to be observable under random sampling but are key to learning good policies. In the future we would like to develop fingerprints for discrete state and action spaces, and explore using a multi-step improvement objective for the BO component.

\section*{Acknowledgements}
We would like to thank Binxin Ru for sharing the code for FITBO, and Yarin Gal for the helpful discussions. This project has received funding from the European Research Council (ERC) under the European Union's Horizon 2020 research and innovation programme (grant agreement \#637713). The experiments were made possible by a generous equipment grant from NVIDIA.

\bibliography{FPO_bib}

\begin{thebibliography}{33}
\providecommand{\natexlab}[1]{#1}
\providecommand{\url}[1]{\texttt{#1}}
\expandafter\ifx\csname urlstyle\endcsname\relax
  \providecommand{\doi}[1]{doi: #1}\else
  \providecommand{\doi}{doi: \begingroup \urlstyle{rm}\Url}\fi

\bibitem[Andrychowicz et~al.(2016)Andrychowicz, Denil, G\'{o}mez, Hoffman,
  Pfau, Schaul, and de~Freitas]{learn_by_grad}
Andrychowicz, M., Denil, M., G\'{o}mez, S., Hoffman, M.~W., Pfau, D., Schaul,
  T., and de~Freitas, N.
\newblock Learning to learn by gradient descent by gradient descent.
\newblock In \emph{Neural Information Processing Systems (NIPS)}, 2016.

\bibitem[Auer et~al.(2002)Auer, Cesa-Bianchi, Freund, and Schapire]{Exp3}
Auer, P., Cesa-Bianchi, N., Freund, Y., and Schapire, R.~E.
\newblock The nonstochastic multiarmed bandit problem.
\newblock \emph{SIAM Journal on Computing}, 32\penalty0 (1):\penalty0 48--77,
  2002.

\bibitem[Brockman et~al.(2016)Brockman, Cheung, Pettersson, Schneider,
  Schulman, Tang, and Zaremba]{OpenAIGym}
Brockman, G., Cheung, V., Pettersson, L., Schneider, J., Schulman, J., Tang,
  J., and Zaremba, W.
\newblock Openai gym, 2016.

\bibitem[Chen et~al.(2017)Chen, Hoffman, Colmenarejo, Denil, Lillicrap,
  Botvinick, and de~Freitas]{learn_without_grad}
Chen, Y., Hoffman, M.~W., Colmenarejo, S.~G., Denil, M., Lillicrap, T.~P.,
  Botvinick, M., and de~Freitas, N.
\newblock Learning to learn without gradient descent by gradient descent.
\newblock In \emph{International Conference on Machine Learning (ICML)}, 2017.

\bibitem[Ciosek \& Whiteson(2017)Ciosek and Whiteson]{OFFER}
Ciosek, K. and Whiteson, S.
\newblock Offer: Off-environment reinforcement learning.
\newblock In \emph{AAAI Conference on Artificial Intelligence}, 2017.

\bibitem[Cox \& John(1992)Cox and John]{cox92sdo}
Cox, D.~D. and John, S.
\newblock A statistical method for global optimization.
\newblock In \emph{IEEE International Conference on Systems, Man and
  Cybernetics}, 1992.

\bibitem[Cox \& John(1997)Cox and John]{cox97sdo}
Cox, D.~D. and John, S.
\newblock \uppercase{SDO}: A statistical method for global optimization.
\newblock In \emph{in Multidisciplinary Design Optimization: State-of-the-Art},
  pp.\  315--329, 1997.

\bibitem[Duan et~al.(2016)Duan, Chen, Houthooft, Schulman, and Abbeel]{rllab}
Duan, Y., Chen, X., Houthooft, R., Schulman, J., and Abbeel, P.
\newblock Benchmarking deep reinforcement learning for continuous control.
\newblock In \emph{International Conference on Machine Learning (ICML)}, 2016.

\bibitem[Finn et~al.(2017)Finn, Abbeel, and Levine]{MAML}
Finn, C., Abbeel, P., and Levine, S.
\newblock Model-agnostic meta-learning for fast adaptation of deep networks.
\newblock In \emph{International Conference on Machine Learning (ICML)}, 2017.

\bibitem[Foerster et~al.(2017)Foerster, Nardelli, Farquhar, Torr, Kohli, and
  Whiteson]{StabExpReplay}
Foerster, J., Nardelli, N., Farquhar, G., Torr, P., Kohli, P., and Whiteson, S.
\newblock Stabilising experience replay for deep multi-agent reinforcement
  learning.
\newblock In \emph{International Conference on Machine Learning (ICML)}, 2017.

\bibitem[Frank et~al.(2008)Frank, Mannor, and Precup]{frank2008}
Frank, J., Mannor, S., and Precup, D.
\newblock Reinforcement learning in the presence of rare events.
\newblock In \emph{International Conference on Machine Learning (ICML)}, 2008.

\bibitem[Glynn(1990)]{Glynn:1990}
Glynn, P.~W.
\newblock Likelihood ratio gradient estimation for stochastic systems.
\newblock In \emph{Communications of the ACM}, 1990.

\bibitem[Graves et~al.(2017)Graves, Bellemare, Menick, Munos, and
  Kavukcuoglu]{Graves}
Graves, A., Bellemare, M.~G., Menick, J., Munos, R., and Kavukcuoglu, K.
\newblock Automated curriculum learning for neural networks.
\newblock In \emph{International Conference on Machine Learning (ICML)}, 2017.

\bibitem[Kakade(2001)]{NPG}
Kakade, S.
\newblock A natural policy gradient.
\newblock In \emph{Neural Information Processing Systems (NIPS)}, 2001.

\bibitem[Krause \& Ong(2011)Krause and Ong]{krause2011contextual}
Krause, A. and Ong, C.~S.
\newblock Contextual gaussian process bandit optimization.
\newblock In \emph{Neural Information Processing Systems (NIPS)}, 2011.

\bibitem[Lillicrap et~al.(2016)Lillicrap, Hunt, Pritzel, Heess, Erez, Tassa,
  Silver, and Wierstra]{DDPG}
Lillicrap, T.~P., Hunt, J.~J., Pritzel, A., Heess, N., Erez, T., Tassa, Y.,
  Silver, D., and Wierstra, D.
\newblock Continuous control with deep reinforcement learning.
\newblock In \emph{International Conference on Learning Representations
  (ICLR)}, 2016.

\bibitem[Malkomes et~al.(2016)Malkomes, Schaff, and Garnett]{Hellinger_kernel}
Malkomes, G., Schaff, C., and Garnett, R.
\newblock Bayesian optimization for automated model selection.
\newblock In \emph{Neural Information Processing Systems (NIPS)}. 2016.

\bibitem[Mordatch et~al.(2015)Mordatch, Lowrey, Andrew, Popovic, and
  Todorov]{Mordatch}
Mordatch, I., Lowrey, K., Andrew, G., Popovic, Z., and Todorov, E.~V.
\newblock Interactive control of diverse complex characters with neural
  networks.
\newblock In \emph{Neural Information Processing Systems (NIPS)}. 2015.

\bibitem[Nichol et~al.(2018)Nichol, Achiam, and Schulman]{Reptile}
Nichol, A., Achiam, J., and Schulman, J.
\newblock On first-order meta-learning algorithms.
\newblock \emph{CoRR}, abs/1803.02999, 2018.

\bibitem[Paul et~al.(2018)Paul, Chatzilygeroudis, Ciosek, Mouret, Osborne, and
  Whiteson]{ALOQ}
Paul, S., Chatzilygeroudis, K., Ciosek, K., Mouret, J.-B., Osborne, M., and
  Whiteson, S.
\newblock Alternating optimisation and quadrature for robust control.
\newblock In \emph{AAAI Conference on Artificial Intelligence}, 2018.

\bibitem[Peters \& Schaal(2006)Peters and Schaal]{PetersRoboticsPG}
Peters, J. and Schaal, S.
\newblock Policy gradient methods for robotics.
\newblock In \emph{2006 IEEE/RSJ International Conference on Intelligent Robots
  and Systems}, 2006.

\bibitem[Pinto et~al.(2017)Pinto, Davidson, Sukthankar, and Gupta]{RARL}
Pinto, L., Davidson, J., Sukthankar, R., and Gupta, A.
\newblock Robust adversarial reinforcement learning.
\newblock In \emph{International Conference on Machine Learning (ICML)}, 2017.

\bibitem[Rajeswaran et~al.(2017{\natexlab{a}})Rajeswaran, Ghotra, Levine, and
  Ravindran]{EPOpt}
Rajeswaran, A., Ghotra, S., Levine, S., and Ravindran, B.
\newblock \uppercase{EPO}pt: Learning robust neural network policies using
  model ensembles.
\newblock \emph{International Conference on Learning Representations (ICLR)},
  2017{\natexlab{a}}.

\bibitem[Rajeswaran et~al.(2017{\natexlab{b}})Rajeswaran, Lowrey, Todorov, and
  Kakade]{linear_policies}
Rajeswaran, A., Lowrey, K., Todorov, E.~V., and Kakade, S.~M.
\newblock Towards generalization and simplicity in continuous control.
\newblock In \emph{Neural Information Processing Systems (NIPS)}.
  2017{\natexlab{b}}.

\bibitem[Rasmussen \& Williams(2005)Rasmussen and Williams]{GPML}
Rasmussen, C.~E. and Williams, C. K.~I.
\newblock \emph{Gaussian Processes for Machine Learning (Adaptive Computation
  and Machine Learning)}.
\newblock The MIT Press, 2005.

\bibitem[Ru et~al.(2017)Ru, McLeod, Granziol, and Osborne]{FITBO}
Ru, B., McLeod, M., Granziol, D., and Osborne, M.~A.
\newblock Fast {Information}-theoretic {Bayesian} {Optimisation}.
\newblock In \emph{International Conference on Machine Learning (ICML)}, 2017.

\bibitem[Schulman et~al.(2015)Schulman, Levine, Abbeel, Jordan, and
  Moritz]{TRPO}
Schulman, J., Levine, S., Abbeel, P., Jordan, M., and Moritz, P.
\newblock Trust region policy optimization.
\newblock In \emph{International Conference on Machine Learning (ICML)}, 2015.

\bibitem[Schulman et~al.(2016)Schulman, Moritz, Levine, Jordan, and
  Abbeel]{GAE}
Schulman, J., Moritz, P., Levine, S., Jordan, M., and Abbeel, P.
\newblock High-dimensional continuous control using generalized advantage
  estimation.
\newblock In \emph{Proceedings of the International Conference on Learning
  Representations (ICLR)}, 2016.

\bibitem[Sutton \& Barto(1998)Sutton and Barto]{Sutton&Barto}
Sutton, R.~S. and Barto, A.~G.
\newblock \emph{Reinforcement Learning : An Introduction}.
\newblock MIT Press, 1998.

\bibitem[Toscano-Palmerin \& Frazier(2018)Toscano-Palmerin and
  Frazier]{BO_expensive_integrands}
Toscano-Palmerin, S. and Frazier, P.~I.
\newblock {Bayesian Optimization with Expensive Integrands}.
\newblock \emph{ArXiv e-prints}, March 2018.

\bibitem[Wang et~al.(2013)Wang, Zoghi, Hutter, Matheson, and
  De~Freitas]{wang_bayesian_2013}
Wang, Z., Zoghi, M., Hutter, F., Matheson, D., and De~Freitas, N.
\newblock Bayesian {Optimization} in {High} {Dimensions} via {Random}
  {Embeddings}.
\newblock In \emph{{IJCAI}}, pp.\  1778--1784, 2013.

\bibitem[Williams et~al.(2000)Williams, Santner, and Notz]{williams_santner}
Williams, B.~J., Santner, T.~J., and Notz, W.~I.
\newblock Sequential design of computer experiments to minimize integrated
  response functions.
\newblock \emph{Statistica Sinica}, 2000.

\bibitem[Williams(1992)]{Reinforce}
Williams, R.~J.
\newblock Simple statistical gradient-following algorithms for connectionist
  reinforcement learning.
\newblock \emph{Machine Learning}, 1992.

\end{thebibliography}
\bibliographystyle{icml2019}
\newpage
\clearpage

\appendix
\section{Appendices}

\subsection{Hyperparameter Settings}
\label{app:hyper_settings}
Our implementation is based on rllab \cite{rllab} and as such most of the hyperparameter settings were kept to their default settings. The details are provided in Table \ref{tab:exp_details}. 
\begin{table}[h]
	\centering
	\caption{Experimental details}
	\begin{tabular}{l c  c  c}
		\toprule
		{}  & Cliff Walker & HalfCheetah & Ant \\
		\midrule
		\PolOpt & TRPO & TRPO & TRPO \\
		KL constraint & 0.01 & 0.01 & 0.01 \\
		Discount rate & 0.99 & 0.99 & 0.99 \\
		GAE $\lambda$ & 1.0 & 1.0 & 1.0 \\
		Batch size & 10,000 & 12,500 & 12,500 \\
		\midrule
		Policy layers & (5,5) & (100,100) & (100,100) \\
		Policy units & Tanh & ReLU & ReLU \\
		\midrule
		FPO-UCB $\kappa$ & 2 & 2 & 2 \\
		\bottomrule
	\end{tabular}
	\label{tab:exp_details}
\end{table}

\subsection{Detailed Experimental Results}
\label{app:quartiles}
In Table \ref{tab:quartiles} we present the quartiles of the expected return of the final learnt policy for each method across the 10 random starts.

\begin{table}
	\centering
	\caption{Quartiles of expected return across 10 random starts}
	\begin{subtable}{1\linewidth}
		\centering
		\vspace{3mm}
		\caption{Cliff Walker}
		\label{tab:cliff_results}
		\begin{tabular}{l c  c  c}
			\toprule
			{}  & Q1 & Median & Q2 \\
			\midrule
			FPO-UCB(S) & 427.1 & 441.5 & 450.0 \\
			FPO-UCB(A) & 335.2 & 432.6 & 440.4 \\
			FPO-FITBO(S) & 428.1 & 443.6 & 453.1 \\
			FPO-FITBO(A) & 372.2 & 438.2 & 451.5 \\
			\midrule
			Na\"ive & -1478.7 & -135.5 & 243 \\
			EPOpt & -44.4 & 282.1 & 354.4 \\
			ALOQ & 33.5 & 57.2 & 77.2 \\
			Random & 345.8 & 358.9 & 373.4 \\
			\bottomrule
		\end{tabular}
	\end{subtable}
	\vspace{3mm}

	\begin{subtable}{1\linewidth}
		\centering
		\caption{HalfCheetah}
		\label{tab:cheetah_results}
		\begin{tabular}{l c  c  c}
			\toprule
			{}  & Q1 & Median & Q2 \\
			\midrule
			FPO-UCB(S) & 3913.7 & 5464.0 & 5905.5 \\
			FPO-UCB(A) & 4435.6 & 5231.8 & 5897.6 \\
			FPO-FITBO(S) & 2973.9 & 3187.2 & 3923.7 \\
			FPO-FITBO(A) & 3686.2 & 4091.1 & 7247.3\\
			\midrule
			Na\"ive & 1059.9 & 1071.1 & 1086.0 \\
			EPOpt & 803.6 & 4066.0 & 4421.0 \\
			OFFER & 1093.9 & 1097.4 & 1111.2 \\
			Random & 1722.3 & 2132.6 & 2645.5 \\
			Enum & 2442.8 & 2796.0 & 3428.4 \\
			\bottomrule
		\end{tabular}
	\end{subtable}
	\vspace{3mm}

	\begin{subtable}{1\linewidth}
		\centering
		\caption{Ant}
		\label{tab:ant_results}
		\begin{tabular}{l c  c  c}
			\toprule
			{}  & Q1 & Median & Q2 \\
			\midrule
			FPO-UCB(S) & 490.6 & 674.2 & 713.3 \\
			FPO-UCB(A) & 408.1 & 519.4 & 629.7 \\
			FPO-FITBO(S) & 626.8 & 704.2 & 770.0 \\
			FPO-FITBO(A) & 455.7 & 533.1 & 707.0 \\
			\midrule
			Na\"ive & -1746.9 & -1669.4 & -1585.2 \\
			EPOpt & -1732.3 & -1606.6 & -1454.5 \\
			Random & 460.3 & 575.6 & 640.4 \\
			Enum & 255.7 & 273.4 & 285.6 \\
			\bottomrule
		\end{tabular}
	\label{tab:quartiles}
	\end{subtable}
\end{table}

\subsection{Further Examination of the Learnt Policies}
\label{app:more_experiments}
\begin{figure*}
	\centering
	\begin{subfigure}{0.45\linewidth}
		\includegraphics[width=1\linewidth]{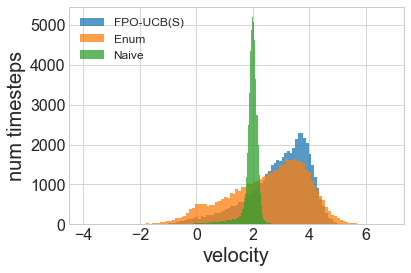}
		\caption{HalfCheetah}
		\label{fig:cheetah_vel}
	\end{subfigure}\hfill
	\begin{subfigure}{0.45\linewidth}
		\includegraphics[width=1\linewidth]{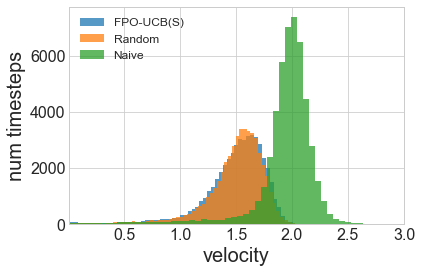}
		\caption{Ant}
		\label{fig:ant_vel}
	\end{subfigure}\hfill
	\caption{Histogram of the velocity profile of the final learnt policies for each method.}
	\label{fig:vels}
\end{figure*}

As explained in the Experiments section, the SREs for the HalfCheetah and Ant experiments are based on the velocities achieved by the agent; for HalfCheetah the SRE is defined as the velocity target being 4 (and carrying a large bonus reward) instead of 2 with a 2\% probability of occurrence, while for Ant velocities greater than 2 has a 5\% probability of incurring a large cost. Here we compare the performance of FPO-UCB(S) against the next best baseline (Enum for HalfCheetah and Random for Ant) and the Na\"ive baseline by visualising the velocity profiles of the final learnt policy. For each random start for each method we sampled 10 trajectories (for a total of 100 trajectories per method) and plot the histogram of the velocity at each timestep. This is presented in Figure \ref{fig:vels}.

For the HalfCheetah task, from Figure \ref{fig:cheetah_vel} we can see that the velocity profile for the Na\"ive approach is highly concentrated around 2. This goes to show that the Na\"ive approach learns a policy that does not take into account the SRE at all. On the other hand, both Enum and FPO-UCB(S) have velocity profiles with much higher variance with a lot of mass spread between 2 and 4. This goes to show that both of them take into account the effect of SREs. However, FPO-UCB(S) manages to better balance the SRE/non-SRE rewards and has slightly higher mass concentrated on 4, which in turn leads to it significantly outperforming Enum.

For the Ant task, unsurprisingly once again the Na\"ive approach completely ignores the SREs, and exhibits a velocity profile that is greater than 2 roughly 50\% of the time. The velocity profiles of the Random baseline and FPO-UCB(S) are almost exactly the same. This is unexpected as there is no significant difference between the expected return of the final policies learnt by these two methods, as shown in Section \ref{sec:experiments}. As noted earlier, the good performance of Random is not unsurprising since the schedule for $\psi$ chosen by FPO-UCB(S) is close to 0.5, which is also the mean of $\psi$ under the random baseline as $\psi \sim U(0,1)$.

\subsection{Performance in settings without SREs}
\label{app:no_sre_cliff}
\begin{figure}
	\centering
	\includegraphics[width=0.9\linewidth]{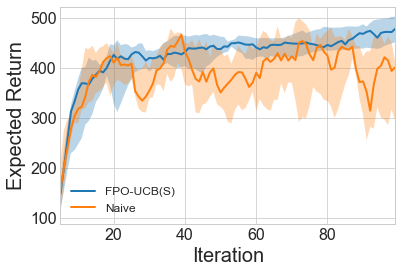}
	\caption{Results for the Cliff Walker environment without any SRE. Solid line shows the median and the shaded region the quartiles across 10 random starts.}
	\label{fig:no_sre_cliff_walker}
\end{figure}

FPO considers the setting where environments are characterised by SREs. A natural question to ask is how does its performance compare to the na\"ive method in settings where there are no SREs. To investigate this we applied FPO-UCB(S), and the na\"ive baseline, to the cliff walker problem presented in Section \ref{sec:exp_cliff_walker}, with the modification that falling off the cliff now carries 0 reward instead of -5000. This removes the SRE, but the environment variable (the location of the cliff) is still relevant since it has a significant effect on the dynamics. The results are presented in Figure \ref{fig:no_sre_cliff_walker}. Note that the performance of the Na\"ive method is far more stable than in the setting with SRE. However, while it is able to learn a good policy, FPO-UCB(S) still performs better since it takes into account the effect of the environment variable.

\end{document}